\documentclass[runningheads]{llncs}
 
\usepackage[mobile]{eccv}

\usepackage{eccvabbrv}

\usepackage{graphicx}
\usepackage{adjustbox}
\usepackage{booktabs}
\usepackage{multirow}
\usepackage{array}

\newcolumntype{P}[1]{>{\centering\arraybackslash}p{#1}}

\usepackage[accsupp]{axessibility}  

\usepackage{hyperref}

\usepackage{orcidlink}

\usepackage{utfsym}
\newcommand{\crossmark}{\scalebox{0.75}{\usym{2613}}}

\begin{document}

\title{Insect Identification in the Wild: \texorpdfstring{\\} TThe AMI Dataset} 

\titlerunning{Insect Identification in the Wild}

\author{Aditya Jain\inst{1}\thanks{Denotes equal contribution. Correspondence to: \url{moth-ai@mila.quebec}}\orcidlink{0009-0009-4630-0651} \and
Fagner Cunha\inst{2\star}\orcidlink{0000-0002-7495-2552} \and 
Michael James Bunsen\inst{1\star}\orcidlink{0000-0002-1458-1382} \and 
Juan Sebastián Cañas\inst{1}\orcidlink{0000-0003-0365-5005} \and
Léonard Pasi\inst{1} \and 
Nathan Pinoy\inst{3} \and
Flemming Helsing\inst{3} \and
JoAnne Russo\inst{4}\orcidlink{0009-0000-6229-5092} \and
Marc Botham\inst{5}\orcidlink{0000-0002-5276-1405} \and
Michael Sabourin\inst{6} \and
Jonathan Fréchette\inst{12} \and
Alexandre Anctil\inst{7} \and
Yacksecari Lopez\inst{8} \and
Eduardo Navarro\inst{8} \and
Filonila Perez Pimentel\inst{8} \and
Ana Cecilia Zamora\inst{8} \and
José Alejandro Ramirez Silva\inst{8}\orcidlink{0000-0001-6799-5207} \and
Jonathan Gagnon\inst{12} \and 
Tom August\inst{5}\orcidlink{0000-0003-1116-3385} \and
Kim Bjerge\inst{3}\orcidlink{0000-0001-6742-9504} \and
Alba Gomez Segura\inst{5}\orcidlink{0000-0002-9575-7262} \and 
Marc Bélisle\inst{9}\orcidlink{0000-0003-0657-6724} \and
Yves Basset\inst{8}\orcidlink{0000-0002-1942-5717} \and
Kent P. McFarland\inst{4}\orcidlink{0000-0001-7809-5503} \and
David Roy\inst{5}\orcidlink{0000-0002-5147-0331} \and
Toke Thomas Høye\inst{3}\orcidlink{0000-0001-5387-3284} \and
Maxim Larrivée\inst{10}\orcidlink{0000-0002-2925-3736} \and 
David Rolnick\inst{1, 11}\orcidlink{0000-0002-2855-393X}}

\authorrunning{A.~Jain et al.}

\institute{Mila - Quebec AI Institute \and
Federal University of Amazonas \and
Aarhus University \and 
Vermont Center for Ecostudies \and
UK Centre for Ecology \& Hydrology \and
McGuire Center \and
Centre de données sur le patrimoine naturel du Québec \and
Smithsonian Tropical Research Institute \and
Université de Sherbrooke \and
Montreal Insectarium \and
McGill University \and
Independent Researcher}

\maketitle

\vspace{-0.3cm}  


\begin{abstract}
Insects represent half of all global biodiversity, yet many of the world's insects are disappearing, with severe implications for ecosystems and agriculture. Despite this crisis, data on insect diversity and abundance remain woefully inadequate, due to the scarcity of human experts and the lack of scalable tools for monitoring. Ecologists have started to adopt camera traps to record and study insects, and have proposed computer vision algorithms as an answer for scalable data processing. However, insect monitoring in the wild poses unique challenges that have not yet been addressed within computer vision, including the combination of long-tailed data, extremely similar classes, and significant distribution shifts. We provide the first large-scale machine learning benchmarks for fine-grained insect recognition, designed to match real-world tasks faced by ecologists. Our contributions include a curated dataset of images from citizen science platforms and museums, and an expert-annotated dataset drawn from automated camera traps across multiple continents, designed to test out-of-distribution generalization under field conditions. We train and evaluate a variety of baseline algorithms and introduce a combination of data augmentation techniques that enhance generalization across geographies and hardware setups. The dataset is made publicly available\footnote[1]{\url{https://github.com/RolnickLab/ami-dataset}}.  

\end{abstract}    
\section{Introduction}
\label{sec:intro}

Computer vision algorithms for species identification have become important tools in studying and protecting the Earth's wildlife. Biodiversity applications have also been beneficial to computer vision, serving as a fruitful source of new benchmarks and algorithms \cite{tuia2022perspectives, beery2021scaling, borowiec2022deep, weinstein2018computer}. However, research in computer vision for biodiversity has often focused on large animals, not reflecting the full set of challenges faced in ecology and conservation.

There are over a million known species of insects on Earth, representing nearly one half of all known organisms (by comparison, there are only 10,000 species of birds, and even fewer mammals) \cite{stork2018many, mora2011many}. Insects play essential roles in nearly all terrestrial ecosystems, as well as providing ecosystem services such as pollination that are critical for humanity \cite{seibold2019arthropod}, yet a looming ``insect apocalypse'' threatens both their abundance and diversity \cite{wagner2020insect}. Despite the crisis, data on insect populations is shockingly sparse \cite{didham2020interpreting}, as a result of less resources, a plethora of species, and the high level of expertise typically required to study insects. Amid this data gap, computer vision tools for automated monitoring have the potential to radically reshape the study and conservation of insects. In particular, recent breakthroughs in camera traps for insects have made it possible to automatically photograph large numbers of individuals in the wild \cite{van2022emerging, hoye2021deep, roy2024}, opening up the the potential for image-processing algorithms to parse this raw data.

Species recognition from camera traps is a well-studied problem in the computer vision community \cite{oliver2023camera}, with common challenges including poor lighting, occlusion of individuals, camouflage, and blur \cite{beery2018recognition}. However, working with insects presents distinct challenges from those faced in traditional camera traps, which are often designed for large animals. For example, while traditional camera trap images may only rarely contain a target species, almost every image from an insect camera trap includes insects -- often hundreds of them. Where traditional camera traps may have a short list of species of interest, insect monitoring requires considering thousands of species at any given location, with many of them being extremely similar, and most with very little labeled data.


\begin{figure}[t!]
    \centering
    \includegraphics[width=1\linewidth]{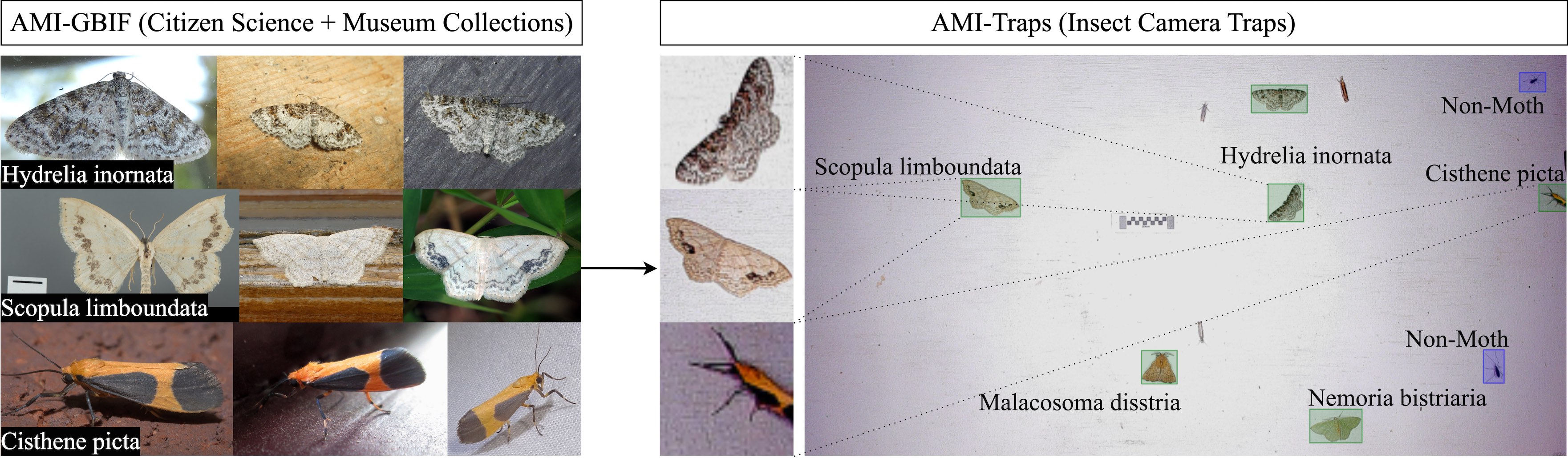}
    \caption{Illustration of the difference between insect species identification from human-captured images and in the wild. On left are shown images of three moth species from our AMI-GBIF dataset, curated from citizen science data and museum collections. On right are shown images of the same species (along with other insects) within a photograph taken by an automated camera trap in the wild from our AMI-Traps dataset. One of the challenges of the AMI dataset is generalizing from AMI-GBIF to AMI-Traps without additional labeled data.}
    \label{fig:what-figure}
\end{figure}


While recent authors have proposed the use of computer vision tools for detecting and identifying insects in camera trap data \cite{bjerge2022real, bjerge2023accurate, suto2022codling, geissmann2022sticky, alison2022moths, korsch2021deep}, such studies have involved small datasets with only a few species represented. There has to date been no work assessing the performance of algorithms in identifying large number of insect species, or addressing the challenges of generalization across in-the-wild deployments. Our work here offers the following contributions:

\begin{itemize}
    \item We introduce the \textbf{AMI} (Automated Monitoring of Insects) dataset, consisting of two parts: 1)  \textbf{AMI-GBIF}, a dataset of $\sim$2.5M human-captured images curated from citizen science platforms and museum collections, 2)  \textbf{AMI-Traps}, an expert-annotated dataset of 2,893 insect camera trap images (representing 52,948 labeled insects) collected from a global network of automated camera traps, designed to test in-the-wild performance (\cref{fig:what-figure}).
    \item We frame the problem of practical insect monitoring in the field as a set of computer vision tasks, entailing very fine-grained classification from long-tailed data, with the additional challenge of generalization across multiple kinds of imagery.
    \item We test a variety of strong baseline models for the proposed tasks, including a combination of data augmentation approaches that significantly improve performance under field conditions.
\end{itemize}

\begin{figure}[t!]
    \centering
    \includegraphics[width=1\linewidth]{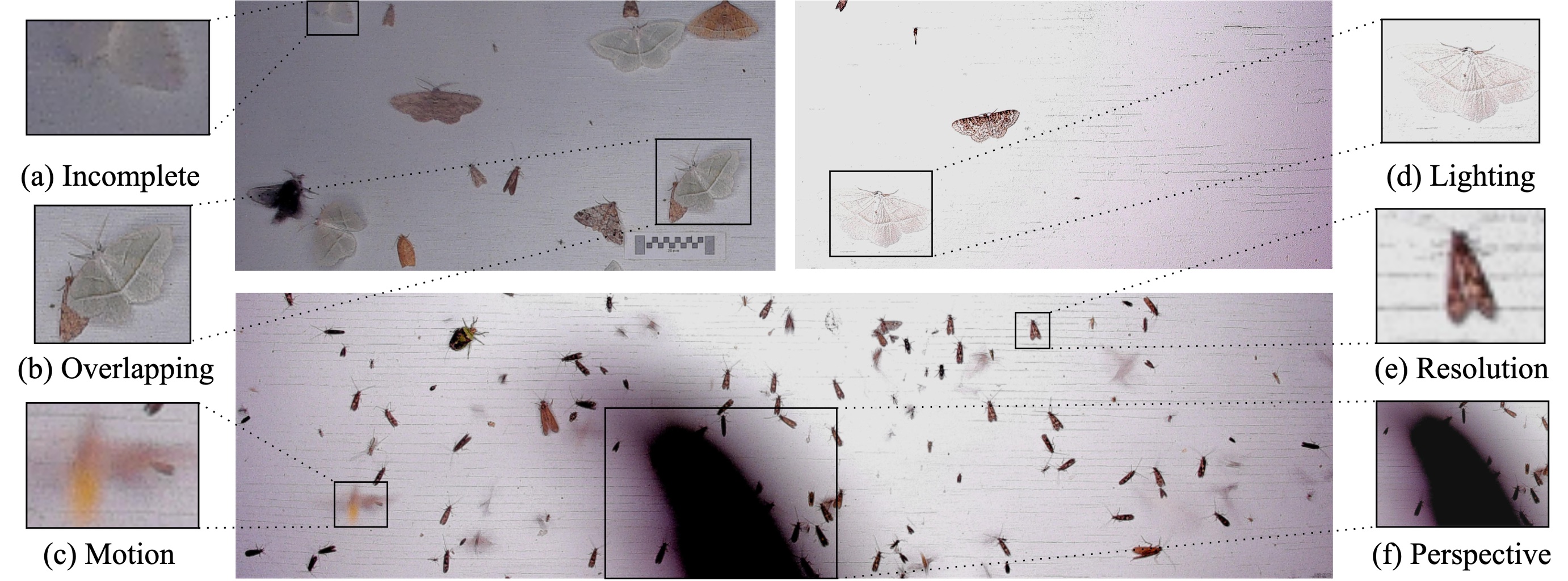}
    \caption{\textbf{Examples of challenges in the AMI-Traps dataset}. \textbf{(a) Incomplete:} In some occasions, the body of an insect is incomplete due to its position in the automatically taken image. \textbf{(b) Overlapping:} Sometimes insects occlude each other. \textbf{(c) Motion:} Moving insects can be blurred and hard to classify. \textbf{(d) Lighting:} The lighting and camera exposure of the images is variable and can lead to poor contrast that masks some or all detail. \textbf{(e) Resolution:} Some insects are very small, leading to low-resolution images. \textbf{(f) Perspective:} While insects are generally positioned at a fixed distance, sometimes insects appear in the air or perched on the camera itself. Additionally, unexpected objects such as spider webs, dirt, and vegetation can be captured by the cameras and may be confused with objects of interest.}
    \label{fig:challenges-ami-traps}
\end{figure}

\section{Related Work}
\label{sec:related_work}

 \textbf{Biodiversity Image Datasets.}  Advancements in computer vision have been closely intertwined with the creation of curated datasets such as ImageNet \cite{deng2009imagenet, russakovsky2015imagenet},  Pascal VOC \cite{everingham2010pascal}, Open Images \cite{krasin2017openimages}, and MS-COCO \cite{lin2014microsoft}. In recent years, the computer vision and machine learning communities have increasingly directed their focus towards addressing pressing human threats, including biodiversity loss \cite{tuia2022perspectives} and climate change \cite{rolnick2022tackling}. These domains not only hold significant potential for societal impact but also offer fertile ground for exploring intriguing machine learning questions such as domain generalization \cite{zhou2022domain}, long-tailed distributions \cite{van2017devil}, and fine-grained classification \cite{wei2021fine}. Such challenges have found especially fruitful testbeds in the classification of species, where datasets such as iNaturalist \cite{van2018inaturalist} have proven useful for both the computer vision and ecology communities. While some challenges have considered plant identification \cite{kumar2012leafsnap,nilsback2008automated,beery2022auto,ouaknine2023openforest}, the majority have focused on animals, including datasets focused on birds \cite{wah2011caltech, berg2014birdsnap, van2015building}, mammals\cite{swanson2015snapshot, anton2018monitoring, beery2018recognition}, and fish \cite{boom2014research, cutter2015automated, kay2022caltech, katija2022fathomnet}. Such datasets have often leveraged concurrent advancements in sensor technology, including in satellite imagery and drones. In particular, camera traps have been widely adopted by biologists as an in situ and cost-effective means of wildlife monitoring \cite{oliver2023camera}, and have in turn presented fascinating challenges for the computer vision community \cite{beery2018recognition}, including in classification \cite{beery2021iwildcam, dubey2021adaptive}, object detection \cite{beery2019efficient, object_detection_camera_trap}, out-of-distribution generalization \cite{ming2022impact, gulrajani2020search, wilds2021}, and active learning \cite{norouzzadeh2021deep}. Such algorithms have to date, however, been designed mostly for large animals such as mammals.

\textbf{Insect Image Datasets.} As algorithms and datasets for species recognition have proliferated, insects have traditionally been underrepresented. Within ImageNet1k \cite{russakovsky2015imagenet}, only 27 classes depicted insects, as compared to 57 for birds and 93 classes for wild mammals 
\cite{luccioni2023bugs}. Some popular fine-grained classification datasets such as NABirds and CUB have focused exclusively on birds \cite{wah2011caltech, van2015building}, while camera trap datasets such as iWildCam have been focused on mammals \cite{beery2021iwildcam}. The iNat2017 dataset \cite{van2018inaturalist} included 1,021 classes of insects out of 5,089 total classes in the dataset. By largely focusing on other taxonomic groups, prior works have neglected the challenges posed by insect data, such as extremely similar classes (reflecting far greater diversity than any other group of organisms), small size, hyperabundance, etc. In recent years, however, some works have introduced insect-focused benchmark datasets. The newly released BIOSCAN-1M dataset includes images and DNA sequences, with a focus on fly identification \cite{gharaee2023step}. Despite the very large size of the dataset (1.1 million datapoints), the number of object classes involved is relatively small, with the focus being to distinguish between 40 taxonomic families of flies, reflecting the challenging nature of insect identification. Other recent datasets for computer vision include one focused on 8 classes of pollinators across Europe \cite{stark2023yolo}, and the Indian Butterflies dataset, encompassing 30k images of butterflies from India across 315 species \cite{indian_butterflies}. A variety of  publications have used computer vision algorithms for detection of pest insects in specific agricultural applications \cite{teixeira2023systematic, badgujar2023real, nawoya2024computer}. 

\textbf{Insect Camera Traps.} The use of camera traps for studying insect biodiversity is relatively recent. Geissmann et al. \cite{geissmann2022sticky} introduced the Sticky Pi device, which uses sticky cards to capture (and kill) insects, which are then automatically photographed. Bjerge et al. \cite{bjerge2021automated} introduced an automated camera trap using UV light to attract and photograph moths and other insects (without killing them), which was built on by subsequent works \cite{bjerge2022real,bjerge2023object}. A similar approach focused on diurnal pollinators via a camera trap installed above flowers \cite{bjerge2023accurate, bjerge2023hierarchical}. These works also propose deep learning methods for processing the resulting images, considering both object detection and classification of a small number of insect categories. The lack of fine-grained classification is largely due to the time required to annotate images from the traps, making it difficult to obtain a large enough labeled dataset from trap data alone.
\section{Benchmark Design}
\label{sec:benchmark_tasks}

In this paper, we propose a benchmark dataset for insect species identification that addresses the challenges faced in real-world use cases. These includes both (1) challenges that distinguish insect identification from other kinds of species identification, notably the massive diversity of often-similar species, and (2) the particular challenges associated with deploying these algorithms ``in the wild''.

To address the first family of challenges, we introduce a dataset for fine-grained insect classification that (to our knowledge) is by far the largest yet released, with $\sim$2.5M images across over 5,000 classes. We concentrate on moths, which are readily drawn to UV camera traps and can often be visually identified to species (unlike some other insects) \cite{bjerge2021automated}.

The second family of challenges is more complex -- involving the gap between field conditions and the high-quality images typically used for training species identification algorithms. Figure \ref{fig:challenges-ami-traps} illustrates some of the challenges that computer vision algorithms face on data from automated insect camera traps, in contrast to images taken by humans. Insects may occlude each other or be partially ``out of frame''. Motion blur and lighting issues may occur, along with objects in unexpected positions. The resolution of images may be very low, owing to the small size of insects at a stereotyped distance from the camera. Also, insect camera traps typically use webcams and PCB camera modules due to their cost and software integration capabilities, and the resulting images are often both structurally different and overall lower quality than those taken by human photographers.

As described in the preceding section, previous authors working with insect camera traps have addressed the problem of challenging camera trap data by training algorithms on annotated data from the particular camera trap in question. However, manual annotation is extremely time-intensive, since it must be performed by expert entomologists, of whom there are very few in any given region. Insect species vary greatly across regions, meaning that fine-grained annotations in one location are generally not useful in another. Furthermore, the hardware used in insect camera traps also varies greatly (with considerable differences in lighting and background, as well as image sharpness, contrast, noise reduction settings, and compression artifacts, which are exacerbated on small scale crops), meaning that even two datasets from the same region may not be comparable.

In this paper, we remove the problem of annotation by solving a more challenging computer vision problem. Namely, we propose to train algorithms on abundant labeled images of insects taken by human photographers, then use these algorithms on out-of-distribution camera trap images. Our benchmark dataset is designed to test this functionality, by including both a large training set of curated photographs taken by humans and an out-of-distribution test set drawn from a geographically diverse set of insect camera traps.

The AMI dataset is therefore designed to test fine-grained classification of moth species both in in-distribution and out-of-distribution contexts. It also includes the task of binary classification of insects as moth or non-moth, which is a prerequisite for camera traps since moths are not the only insects to appear. Note that we do not consider the task of object detection, which in applications is a precursor to insect identification. This is because object detection against a relatively plain  background is a simpler task, where good results have already been obtained in the insect camera trap literature \cite{bjerge2021automated,bjerge2022real,bjerge2023object,geissmann2022sticky}.
\section{The AMI Dataset}
\label{sec:dataset_creation}
\begin{figure}[t!]
    \centering
    \includegraphics[width=\textwidth]{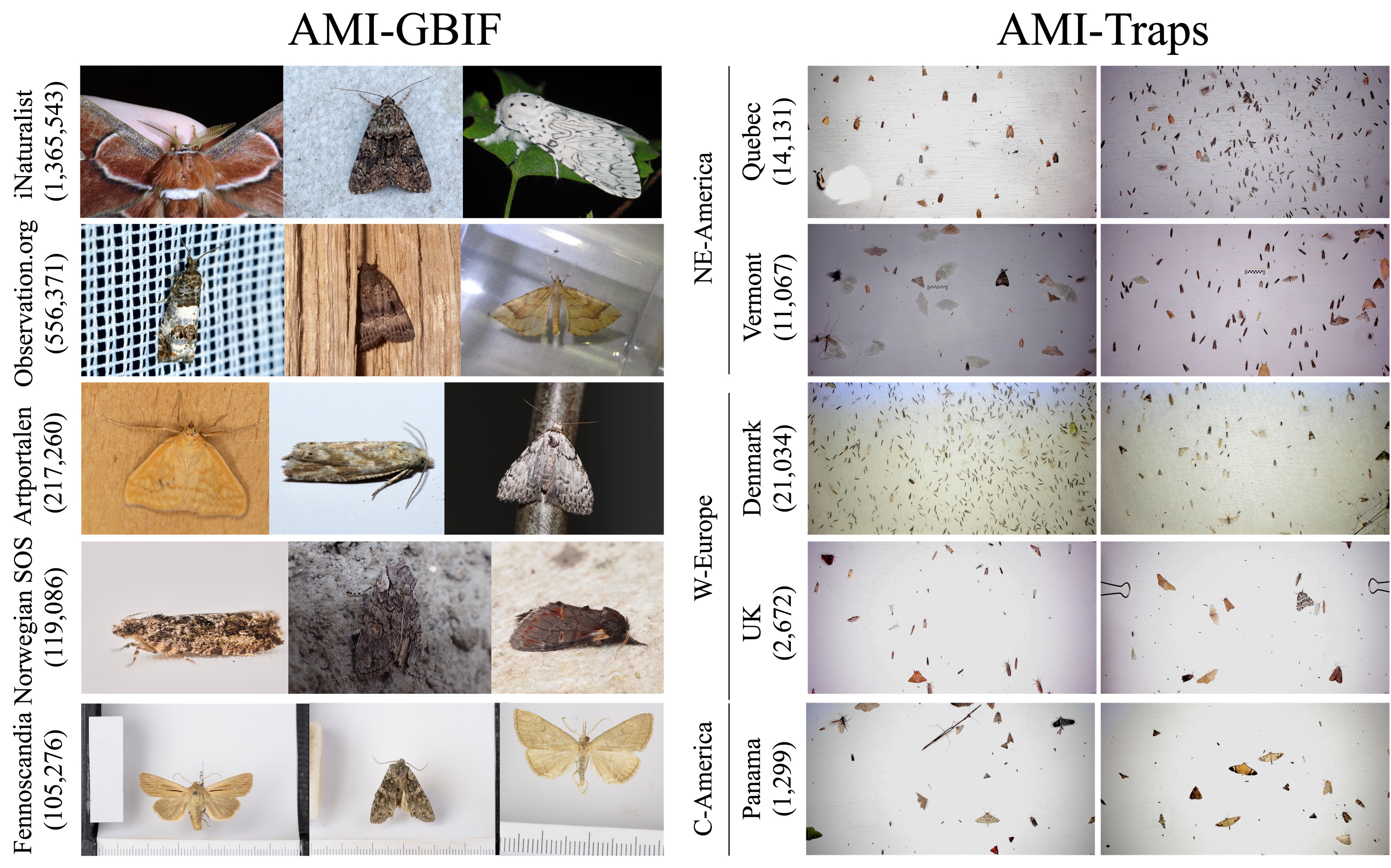}
    \caption{\textbf{Sample images from the AMI Dataset.} The AMI Dataset is composed of (1) AMI-GBIF, curated from a number of sources with imagery from citizen science and museum collections, and (2) AMI-Traps, drawn from automated camera traps for insects across five countries in three regions. The number of individual insects annotated per sub-dataset is denoted in parentheses.}
    \label{fig:overview}
\end{figure}

In this section, we describe the construction of the datasets that we release in this work: \textbf{AMI-Traps} and \textbf{AMI-GBIF}. For AMI-Traps, we collected insect camera trap data by field deployments in three regions: NE-America (Northeast of North America), which includes data from Quebec in Canada and Vermont in the USA; W-Europe (Western Europe), which includes data from the UK and Denmark; and C-America (Central America), comprising data from Panama. For AMI-GBIF, we compiled public data for the same three regions using the Global Biodiversity Information Facility (GBIF)~\cite{gbif2023}. The GBIF platform provides access to biodiversity data aggregated from multiple sources, such as museums, research institutions, and citizen science platforms such as iNaturalist~\cite{inat2024}, Observation.org~\cite{obs2024}, and BugGuide~\cite{bug2024}. For brevity, we sometimes refer to AMI-Traps and AMI-GBIF as AMI-T and AMI-G, respectively. \Cref{fig:overview} shows examples of images from the two datasets and \cref{tab:comparison_ami-gbif_ami-traps} compares the number of images and coverage of classes at different taxonomic levels. 

\subsection{AMI-Traps Dataset}\label{subsec:ami-traps}

\textbf{Data Collection}. The source images in the AMI-Traps dataset were collected from a variety of different insect camera traps deployed across 22 monitoring stations in the three regions previously described. The cameras were configured to be motion-sensitive, as well as taking photos on a fixed interval of 10 minutes; we selected a subset of images for annotation so as to avoid highly similar images. In total, 2,893 trap images were sampled from all cameras and made available on the Label Studio annotation platform (see \cref{sec:ami-traps-construction-appendix} for more details).

\textbf{Annotation Process}. Images were annotated by entomologists with expertise in the species of the particular region. Labeling was conducted according a three-step process. 1) We designed a custom object detection model to suggest boxes around objects in the image (refer to \cref{sec:ami-traps-construction-appendix} for details on this model), to save annotators time. Annotators were asked to correct errors in the proposed boxes as well as drawing any missing boxes. Insects appearing smaller than 1 centimeter were ignored. 2) Annotators were asked to label image crops according to one of three coarse categories: moth, non-moth, or unidentifiable. Some insects could not be labeled due to motion blur, occlusion, or other factors. 3) For insects labeled as moths, annotators were asked to select the narrowest taxonomic group that could be determined from the image - identifying moths to species where possible, if not species then genus, and if not genus then family. 

\textbf{Data Statistics.} 2,893 source images from the traps were annotated (see \cref{tab:source_images} for further details) which includes Vermont (900), Denmark (892), Quebec (533),  UK (446), and Panama (122). A total of 52,948 insects were labeled within these images, including 37,105 (70.08\%) non-moths, 14,105 (26.64\%) moths, and 1,738 (3.28\%) unidentifiable. Out of the 14,105 moth crops, 10,854 (76.95\%), 7,065 (50.09\%) and 5,374 (38.10\%) crops were labeled at the family, genus, and species level respectively.

\subsection{AMI-GBIF Dataset}\label{subsec:ami-gbif}

AMI-GBIF has two sub-datasets: binary and fine-grained classification. The fine-grained classification dataset contains images for moth species in each of the regions covered by AMI-Traps. The binary classification dataset contains images of moths aggregated across all regions and non-moths from anywhere on earth.

\textbf{Dataset Construction}. First, in collaboration with regional moth specialists, we created ``checklists'' of species that could occur in each of the three regions of interest. It is worth noting that some species are included in more than one checklist, though most are unique to a single region. Next, we compared the checklists against the GBIF taxonomic backbone~\cite{gbif_backbone}, merged synonym names (since some species have multiple accepted names), and manually investigated doubtful or unmatched names. Following that, we used the GBIF occurrence search tool to download metadata for all observations from the order Lepidoptera with images, which returned 15.1 million observations~\cite{gbif_leps}. We then fetched images from observations in which the species is included in our regional checklists. The number of training images per species were limited to 1,000, though for many species it was considerably lower due to the long-tailed nature of the data. All images were resized to have the smallest dimension of 450px and packaged using the WebDataset format \cite{WebDataset}.

\textbf{Quality Control}. After downloading the images, we applied a set of cleaning steps to ensure the quality of the dataset. First, we removed duplicate URLs, i.e., images linked to more than one observation. This may occur when an image contains multiple individual insects, or when a placeholder is used for an observation (for example, \cref{fig:gbif_duplicate_placeholder} in \cref{sec:cleaning_steps_ami_gbif} shows a picture used as a placeholder for more than 100,000 observations). Next, we excluded images from specific datasets (refer to ~\cref{table:removed_sources} in \cref{sec:cleaning_steps_ami_gbif}) that we found to have a high percentage of images unsuitable for training our models, for instance, images of body parts or text only. Another issue we identified is that some observations contained thumbnail images that are too small to provide sufficient information to train our models; we removed all images with a width or height less than 64 pixels. Finally, as our dataset targets only adult insects, we removed images containing non-adult individuals such as eggs and caterpillars. For observations where the life stage was reported by the observer, approximately 60\% of the cases, we removed all occurrences tagged as non-adult. For the remaining images with a missing life stage tag, we designed a classification model to predict the life stage (refer to \cref{sec:cleaning_steps_ami_gbif} for details about this model) and removed all images classified as non-adult.

\textbf{Dataset Split}. Each GBIF observation may contain more than one image, but each image is considered an independent instance of the species. However, as images belonging to the same observation may be very similar, we keep images from the same observation in the same data split. To do so, we split the dataset using the observation ID as reference, with the proportion of 70\%, 10\%, and 20\% for the train, validation and test sets respectively. We apply a stratified split, i.e., we guarantee that this proportion is also maintained at the species level. We exclude species with fewer than 5 images, so the minimum number of images for a species in the training set is 4, and in the test set is one. Due to the data split based on observation ID, the validation set might lack images for some species. As some species may belong to multiple checklists, we keep the train/val/test splits consistent across the regions to prevent data leakage when training a unified model for all regions.

\textbf{Binary Classification}. Since moths are not the only insects that appear on the camera trap screen, for our proposed benchmark task of binary classification, we curate a dataset comprising 350,000 images for each of the moth and non-moth categories. The moth class consists of adult moth images randomly sampled from the AMI-GBIF dataset. For the non-moth category, we leveraged expert knowledge of the taxonomic groups likely to appear on the screen, fetching images of adult individuals from GBIF of the following groups: Diptera (85,474), Hemiptera (75,053), Odonata (68,770), Coleoptera (45,954), Araneae (21,321), Orthoptera (21,261), Formicidae (12,206), Ichneumonidae (6,581), Trichoptera (5,316), Neuroptera (4,880), Opiliones (1,880), Ephemeroptera (1,304) --- this includes mostly insects, with some Arachnids. The dataset is split into train, validation, and test sets with proportions of 70\%, 10\%, and 20\%, respectively. Specifically for the moth category, the splits are kept consistent with the splits for the fine-grained species classification task.

\textbf{Statistics}. The final dataset contains $\sim$2.5M images from 5,364 species, 1,734 genera and 77 families of moths, along with 350,000 additional images of non-moth groups. The binary dataset contains 700,000 images, composed of 350,000 moth images and the 350,000 non-moth images. The fine-grained dataset contains 2,564,392 moth images, and \cref{tab:comparison_ami-gbif_ami-traps} shows the number of images and species for each region. Though species are selected based on their presence in regions of interest, we include all GBIF observations of these species from anywhere on Earth (see Fig.~\ref{fig:gbif_global_dist} for the global distribution of images). As in other biodiversity datasets, the number of images per species in AMI-GBIF follows a long-tailed distribution (see Fig.~\ref{fig:long_tailed_distribution}). Specifically, only 3,099 species have more than 100 images and are considered under the ``many-shot'' bucket, 1,358 species contains between 20 and 100 images and are considered ``medium-shot'', and there are 907 species with less than 20 training images, considered as ``few-shot'' categories.

\begin{table}[t!]
    \caption{Comparison between the AMI-GBIF and AMI-Traps dataset for the fine-grained classification task. Images for the traps dataset denote the individual insect crops and not the raw trap images. The last three rows show the unique classes for each of the three taxonomic levels for the two datasets for different regions. It is also important to note that regions have species in common.}
    \label{tab:comparison_ami-gbif_ami-traps}
    \centering
    \begin{tabular}{|c|cc|cc|cc|cc|} 
     \hline
         &  \multicolumn{2}{c}{All Regions}&  \multicolumn{2}{|c|}{NE-America}&  \multicolumn{2}{c|}{W-Europe}&  \multicolumn{2}{c|}{C-America}\\ 
         &  AMI-G&  AMI-T&  AMI-G&  AMI-T&  AMI-G&  AMI-T&  AMI-G& AMI-T\\ 
         \hline
         Images&  2,564,392 &  14,105&  1,179,943 &  9,066&  1,579,333 &  4,066&  99,405& 973\\ 
         Families&  77 &  43&  68&  34&  65 &  26&  27 & 11\\ 
         Genera&  1,734 &  344&  887&  204&  957 &  192&  385 & 12\\
         Species& 5,364 & 516& 2,497& 277& 2,603& 244& 636&6\\ 
    \hline
    \end{tabular}    
\end{table}

\section{Methods}
\label{sec:methods}

In this section, we describe the methods used to assess the difficulty of the proposed benchmark tasks. We evaluate a variety of high-performing image classification architectures and use extensive data augmentation to address the challenges of domain shift between the two datasets. Motivated by the particular structure of the AMI-Traps dataset, we add a mixed-resolution (MixRes) data augmentation technique. As we describe in the following section, our approaches perform quite well, with MixRes leading to an especially significant performance improvement. 

\textbf{Models}. We work with five computer vision models widely used in identification of wildlife: ResNet50~\cite{he2016deep}, MobileNetV3-Large~\cite{howard2019searching}, ConvNeXt-T~\cite{liu2022convnet}, ConvNeXt-B~\cite{liu2022convnet}, and ViT-B/16~\cite{dosovitskiy2020image}. We select ResNet50 because it is the most popular backbone used as a baseline for almost any computer vision task. Considering the possibility that the trained models may need to run on hardware with limited resources, such as edge devices and personal computers, for in-field predictions, we choose MobileNetV3Large as a lightweight model alternative. Furthermore, considering that vision transformers have surpassed convolutional neural networks and become the state-of-the-art models for image classification, we select the ViT-B/16 model to provide a baseline from this family of architectures. Finally, for representing modern convolutional neural networks, we choose the recently introduced ConvNeXt architecture, which has proven to be competitive with transformers using only standard ConvNet modules. From this family of models, we select two versions: the base version (ConvNeXt-B) and, considering lightweight usage scenarios, we also include the tiny version (ConvNeXt-T). Unless otherwise stated, we use a default input resolution of 128x128 for all models.

\textbf{Data Augmentation Approaches}. One characteristic of the in-the-wild images is the large variety of crops due to the diversity of insect sizes. A considerable number of them have small resolutions, below 64 $\times$ 64. At this scale, some important visual features that the model may have learned from high-resolution GBIF images might not be identifiable, leading to inaccurate predictions. This issue can be further amplified when these low-resolution crops are scaled up to the input model size. To compel the model to learn features for low-resolution inputs and avoid bias towards visual features only distinguishable in high-resolution images from other domains, we use a data augmentation technique we refer to as Mixed Resolution (MixRes) that simulates this low-resolution input during training (see Fig.~\ref{fig:mixed-resolution} in supplementary). This approach consists of randomly resizing the training input image to 25\% or 50\% of the model input size with a probability $\rho$ for each case. In our experiments $\rho$ is set to $0.25$, maintaining the original input size with probability $0.5$. We employ MixRes in all experiments unless otherwise stated.

In addition to MixRes, we also use a set of standard data augmentation operations: random resize crop with scale of 0.3 to 1.0 and an aspect ratio between 3/4 and 4/3, random horizontal flip with 50\% probability, and RandAugment~\cite{cubuk2020randaugment} with $N=2$ operations and magnitude $M=9$. Finally, the image is normalized using the mean and standard deviation of ImageNet and scaled to the input network size.

\textbf{Implementation Details}. We train all models with the AMI-GBIF training sets using the ImageNet-1K pretrained weights from the timm library~\cite{rw2019timm} (for ConvNext and ViT we used ImageNet-22K weights). Each model is trained with a batch size of 128 and the AdamW optimizer~\cite{loshchilov2018decoupled} for 30 epochs. The initial learning rate is set to 0.001, with a weight decay of 1e-5. The learning rate is linearly warmed up from 0 to the initial value for 2 epochs and then decayed using cosine scheduling~\cite{he2019bag}. We also apply label smoothing~\cite{szegedy2016rethinking} with a value of 0.1.

\textbf{Inference Settings}. At test time, the insect crops from AMI-Traps are padded with black pixels to make them square, preventing the distortion of the shape of the insect before prediction. Next, the images are normalized with the mean and standard deviation of ImageNet and, finally, resized to the model input size. No padding is done for the test images from AMI-GBIF.

\textbf{Higher Taxonomic Rank Prediction}. We consider the accuracy of our baseline models both with respect to the exact insect species predicted and with respect to higher level taxonomic groups such as genus and family.  Indeed some images in our dataset cannot be labeled at the species level at all due to the quality or perspective of the image, and our annotators instead provided labels at the genus or family level. In our experiments, to make predictions at a higher taxonomic level, we simply sum together species-level predictions to get genus results, and similarly, the genus predictions are grouped together to get family predictions.

\textbf{Evaluation Metrics}. During the binary classification step, it is important to achieve high precision by filtering out as many non-moths as possible to prevent the propagation of errors in the pipeline. At the same time, it is also important to assess recall because if too many moths are discarded at this step, it may impact the utility of the data for ecological studies. Therefore, for the binary classification task, along with accuracy, we also consider precision, recall, and F1-score as evaluation metrics. For the fine-grained classification task, we evaluate the model in terms of micro- and macro-averaged accuracy at the species, genus and family levels. Micro-averaging aggregates all class samples to calculate the accuracy while the macro-average is the average of per-class accuracy. Unless otherwise stated, the reported accuracy is micro-averaged. We conduct each experiment for three independent runs and report the mean and standard deviation.
\section{Experiments and Results}
\label{sec:experiments}

\textbf{Binary Classification Results}. This task involves distinguishing moths from non-moths. Here we train a ResNet50 model on the AMI-GBIF dataset and evaluate performance both on the AMI-GBIF held-out test set and on the entirety of the AMI-Traps dataset. On AMI-GBIF (in-distribution), the model achieves a test accuracy of $97.77\%\pm0.02\%$. \Cref{tab:binary_eval_general} shows the results for the AMI-Traps dataset in more detail. We observe that the model has a high recall but a lower precision, which means it is over-predicting the moth class. Also, the bigger the image crop, the better the accuracy of the insect classification; many non-moths where the model errs are extremely small and are therefore highly blurred in the trap images. \Cref{tab:binary_eval_reshaping} in the supplementary illustrates the improvement in model performance when crops are square padded (our default at inference time) versus an ablation where padding is not performed.

\begin{table}
    \centering
    \caption{Binary classification model evaluation results on the AMI-Traps dataset after training on AMI-GBIF, showing that recall is high throughout, while precision drops on very small insects, which are predominantly non-moths. Crops >100px or >150px represent that atleast one side of the box is greater than the threshold.
    }
    \label{tab:binary_eval_general}
    \resizebox{\textwidth}{!}{%
    \begin{tabular}{|l|c c c P{1.8cm}P{1.8cm}P{1.8cm}P{1.8cm}|}
         \hline
         &  Moths&  Non-moths &Total&  Accuracy&  Precision& Recall& F1-score\\
         \hline
         All crops&  14,105&  37,105 &51,120&  86.48$\pm$1.00&  68.31$\pm$1.70&  95.03$\pm$0.39& 79.48$\pm$1.27\\
         Crops > 100px&  11,630&  16,534 &28,164&  88.39$\pm$0.97&  79.46$\pm$1.43 &  96.99$\pm$0.20 & 87.35$\pm$0.94\\
         Crops > 150px&  6,446&  5,961 &12,407&  92.95$\pm$0.22 &  89.83$\pm$0.43 &  97.48$\pm$0.12 & 93.70$\pm$0.30\\
         \hline
    \end{tabular}}
\end{table}

\textbf{Fine-grained Classification Results}. The fine-grained classification task aims to identify moths to the species level (and secondarily to genus or family level). In our experiments, we train models on AMI-GBIF data for each of the three regions of the AMI dataset, evaluating both on AMI-GBIF (in-distribution) and AMI-Traps (out-of-distribution) data (see \cref{tab:evaluation-results}). Not all moth species present in a region appear regularly at camera traps; hence, the list of species in AMI-Traps is a subset of species in AMI-GBIF for a given area (see \cref{tab:comparison_ami-gbif_ami-traps}). Therefore, we also evaluate the models on an in-distribution subset (Test\ddag) from AMI-GBIF containing only the species present in AMI-Traps. Meanwhile, for AMI-Traps, there are 16 species not present in AMI-GBIF that were removed during our evaluation procedure (these represent unexpected species not included in the regional checklists used to define AMI-GBIF). Overall, the best model at the species level is ConvNeXt-B for both AMI-GBIF and AMI-traps. 
For C-America, annotations are mostly provided at the genus or family level, due to the challenging nature of visual identification of moths in this biodiversity-rich region. Furthermore, while the genus and family accuracies are similar between N-America and W-Europe, they are much lower for C-America. This can be attributed to the fact that there are 10 times fewer images in the training set for this region.
Note that for genus and family evaluation, there are some cases where the accuracy is lower than the species level. This occurs because those higher taxonomic levels include crops that have labels only at higher levels. In the supplementary materials, we include additional results for AMI-GBIF long-tail accuracies (\cref{tab:ami-gbif-long-tail-results}), a pre-training ablation study (\cref{tab:pretraining-abla}), and a study analyzing classification accuracy as a function of prediction confidence (\cref{fig:performance_top1_accuracy}).


\begin{table}[t!]
\centering
\caption{Accuracy results for the fine-grained classification task. Overall, ConvNeXt-B achieves the best results for both AMI-GBIF and AMI-traps, with ViT-B/16 achieving slightly better results for some cases at the genus and family levels. Test$\ddag$ represents a subset of the AMI-GBIF test set, containing only the species present in AMI-Traps. Species-level results on AMI-Traps for the C-America region are not available since most annotations were provided only at higher taxonomic levels.}
\label{tab:evaluation-results}
\resizebox{\textwidth}{!}{%
\begin{tabular}{|l|l|*{4}{P{1.73cm}P{1.73cm}|}}
\hline
\multicolumn{1}{|c|}{\multirow{3}{*}{\rotatebox[origin=c]{90}{Region}}} &
  \multicolumn{1}{c|}{\multirow{3}{*}{Model}} & \multicolumn{2}{c|}{AMI-GBIF} & \multicolumn{6}{c|}{AMI-Traps} \\ \cline{3-10} 
 &
   &
  \multicolumn{2}{c|}{Species} &
  \multicolumn{2}{c|}{Species} &
  \multicolumn{2}{c|}{Genus} &
  \multicolumn{2}{c|}{Family} \\
 &
   &
  \multicolumn{1}{c}{Micro Test} &
  \multicolumn{1}{c|}{Micro Test$\ddag$} &
  \multicolumn{1}{c}{Micro Top1} &
  \multicolumn{1}{c|}{Macro Top1} &
   \multicolumn{1}{c}{Micro Top1} &
  \multicolumn{1}{c|}{Macro Top1} &
   \multicolumn{1}{c}{Micro Top1} &
  \multicolumn{1}{c|}{Macro Top1} \\ \hline
\multirow{5}{*}{\rotatebox[origin=c]{90}{NE-America}} & ResNet50         &     87.56$\pm$0.12          &    89.39$\pm$0.14           & 71.86$\pm$1.02 &  64.96$\pm$2.00 & 80.07$\pm$0.49 &  74.66$\pm$0.35 & 90.44$\pm$0.41 & 69.30$\pm$4.15  \\
                            & MBNetV3L &   83.73$\pm$0.17            &      85.80$\pm$0.29        &  67.41$\pm$0.72 &  61.21$\pm$1.20 & 74.83$\pm$1.30 & 71.20$\pm$2.47 & 89.31$\pm$0.62 &  64.48$\pm$1.87  \\
                            & ConvNeXt-T       &      89.81$\pm$0.05         &     91.21$\pm$0.06          &   74.89$\pm$0.05 & 68.00$\pm$0.76 & 83.26$\pm$1.22 & 77.26$\pm$0.85 & 90.61$\pm$0.53 &  74.60$\pm$0.64  \\
                            & ConvNeXt-B       &      \textbf{90.45$\pm$0.03}         &     \textbf{91.66$\pm$0.03}   &  76.75$\pm$1.52 & \textbf{68.56$\pm$0.62} & 85.46$\pm$0.83 & 78.28$\pm$0.14  & 91.07$\pm$0.29 & \textbf{76.87$\pm$1.10} \\
                            & ViT-B/16         &     86.39$\pm$0.05          &     87.94$\pm$0.10          &  \textbf{77.43$\pm$0.79} & 68.02$\pm$0.34 & \textbf{85.57$\pm$0.98} &  \textbf{78.97$\pm$0.72}  & \textbf{91.45$\pm$0.34} & 73.66$\pm$1.49 \\ \hline
\multirow{5}{*}{\rotatebox[origin=c]{90}{W-Europe}}   & ResNet50         &    86.45$\pm$0.08          &    90.25$\pm$0.03           & 79.39$\pm$0.39 &  72.84$\pm$0.21 & 78.95$\pm$0.26 & 73.22$\pm$0.60 & 89.13$\pm$0.42 & 75.24$\pm$1.85 \\
                            & MBNetV3L &          82.20$\pm$0.05     &      86.47$\pm$0.13        &  76.30$\pm$1.29 & 70.35$\pm$0.93  & 76.35$\pm$1.19 & 72.88$\pm$1.28  & 88.33$\pm$0.47 &  72.21$\pm$0.38 \\
                            & ConvNeXt-T       &     88.82$\pm$0.03          &    92.05$\pm$0.05          &  82.44$\pm$0.89 & 74.59$\pm$0.62  & 81.86$\pm$0.79 &  74.15$\pm$0.59 & 90.08$\pm$0.59 & 74.58$\pm$2.93   \\
                            & ConvNeXt-B       &      \textbf{89.35$\pm$0.01}         &    \textbf{92.48$\pm$0.05} &  \textbf{82.87$\pm$0.39} & \textbf{76.41$\pm$0.50} & 82.45$\pm$0.44 &  \textbf{75.98$\pm$1.05} & 91.27$\pm$0.60 &  \textbf{79.39$\pm$2.70}  \\
                            & ViT-B/16         &      85.45$\pm$0.03         &     89.22$\pm$0.11          &   82.44$\pm$0.36 & 75.25$\pm$0.98 & \textbf{83.01$\pm$0.26} & 75.96$\pm$0.35  & \textbf{91.68$\pm$0.17} & 75.81$\pm$0.79  \\ \hline
\multirow{5}{*}{\rotatebox[origin=c]{90}{C-America}}  & ResNet50         &   87.46$\pm$0.23          &     95.92$\pm$0.68          & - &   - & 47.44$\pm$1.11 & 66.59$\pm$0.60 & 69.18$\pm$0.24 & 42.40$\pm$0.31\\
                            & MBNetV3L &     83.68$\pm$0.24          &    92.89$\pm$0.27            & -  &  -  &  41.03$\pm$2.22 &  59.32$\pm$1.18 & 68.34$\pm$1.69 & 41.42$\pm$2.98  \\
                            & ConvNeXt-T       &     87.65$\pm$0.11          &    95.46$\pm$0.46           &  -  &  -  &  \textbf{55.77$\pm$3.33} & \textbf{70.68$\pm$1.58} & 69.46$\pm$0.42 & \textbf{45.35$\pm$0.67}  \\
                            & ConvNeXt-B       &     \textbf{88.62$\pm$0.09}          &    \textbf{96.22$\pm$0.26}        &  -  & - &    51.92$\pm$3.85 & 68.63$\pm$1.94 & \textbf{72.24$\pm$1.47} & 44.31$\pm$1.32 \\
                            & ViT-B/16       &      79.80$\pm$0.26         &    88.36$\pm$1.02            &  -  & -  &  47.44$\pm$4.44 & 64.70$\pm$4.07 & 69.87$\pm$1.51 & 42.00$\pm$3.60  \\ \hline
\end{tabular}}
\end{table}

\textbf{Data augmentation approaches}. To assess the effectiveness of the mixed resolution data augmentation technique, we conduct an ablation study using ResNet50 for the NE-America region. As the effects of MixRes could already be covered by other data augmentation operations used in our preprocessing, such as random resize crop or one of the operations of RandAugment, we analyze multiple combinations of these operations. Additionally, we evaluate a higher input resolution, using the original 224 input size of ResNet50, to determine whether this leads to better predictions. \Cref{tab:finegrained_ablation} shows the results. While applying MixRes results in a slight decrease in accuracy within the in-distribution AMI-GBIF test set, for the in-the-wild AMI-Traps test set there is a substantial improvement, ranging between 4\% and 7\% when compared to models not trained with MixRes. Another important observation is that, although RandAugment and random resize together contribute to an overall improvement of around 16\%, MixRes remains additionally beneficial. Finally, despite an increase of more than 4\% in accuracy within the original domain when using a higher resolution, there is a marginal decrease in the accuracy of the in-the-wild results.

\begin{table}[t!]
\caption{Ablation study on data augmentation techniques using a ResNet50 for the NE-America region evaluated on micro-averaged top-1 species accuracy. ``RandAug'' denotes the combination of the RandAugment and random resize crop data augmentation techniques. AMI-GBIF$\ddag$ test set contains only the species present in AMI-Traps.}
\label{tab:finegrained_ablation}
\begin{center}
\begin{tabular}{|cccP{1.8cm}P{1.8cm}|}
\hline
 Resolution & RandAug & MixRes & AMI-GBIF$\ddag$ & AMI-Traps \\ \hline
 128 & \crossmark & \crossmark & 89.18$\pm$0.08 &  51.46$\pm$2.73\\
 128 & \checkmark & \crossmark & 89.71$\pm$0.16 &  67.88$\pm$1.27\\
 128 & \crossmark & \checkmark & 87.41$\pm$0.06 &  58.41$\pm$0.30\\
 128 & \checkmark & \checkmark & 89.39$\pm$0.14 &  \textbf{71.86$\pm$1.02}\\
 224 & \checkmark & \checkmark & \textbf{92.97$\pm$0.17} &  68.41$\pm$1.06\\ 
 \hline
\end{tabular}
\end{center}
\end{table}

\section{Conclusion}
\label{sec:conclusion}

Many ecological applications require species-level image classification to meet their goals. Despite advances in computer vision, there remain challenges in implementing fine-grained classification that is effective in real-world scenarios. Here we outline the challenges of insect identification in the wild and propose a benchmark dataset to tackle them. We present AMI-GBIF, a dataset with 2.5 million labeled photographs of 5,000 moth species. We also introduce AMI-Traps, a dataset of images captured by automated insect monitoring systems across five countries, labeled by regional experts. We evaluate multiple classification models both on the AMI-GBIF dataset and in out-of-distribution generalization to AMI-Traps, with ConvNeXt-B generally performing the best. We also evaluate a range of data augmentation techniques, finding that a combination of simple approaches, notably mixed-resolution augmentation, considerably improves generalization under field conditions.

Several limitations of this work should be noted. First, while annotating AMI-Traps already occupied a considerable time for expert entomologists, further expanding it to a greater number of species and geographic regions would be helpful in further stress-testing field deployments of fine-grained classification algorithms. Second, while we focus on camera traps for nocturnal insects, it would be useful also to consider diurnal insect camera traps, for example those aimed at pollinators, which pose some additional challenges, such as multiple perspectives and a diversity of backgrounds. We hope furthermore that future works will consider more sophisticated domain adaptation approaches building on the baselines we present here.

This work represents a step towards computer vision algorithms for species recognition that are effective in the ultra-challenging regime of insect identification and work effectively in field conditions. The approach we propose should make it possible in particular to generalize algorithms for insect camera traps rapidly and cost-effectively across different hardware setups and regions of interest. We hope that the datasets, challenges and strategies presented here will propel further research in computer vision for insects and help accelerate initiatives to understand and protect our planet's biodiversity.
\section*{Acknowledgements}

We thank Annie-Shan Morin, Anna Viklund, and Jamie Alison for their helpful input and support to the project.

This research was generously funded by Mila~-~Quebec AI Institute and Ministère de l’Économie, de l'Innovation et de l'Énergie. The work was additionally supported in part by the Canada CIFAR AI Chairs program (NSERC 585136), the Global Climate Center on AI and Biodiversity Change (ABC) (NSF OISE-2330423), the Montreal Insectarium, the Coordenação de Aperfeiçoamento de Pessoal de Nível Superior - Brasil (CAPES-PROEX) - Finance Code 001, the Amazonas State Research Support Foundation - FAPEAM - through the POSGRAD project 2024/2025, the AMBER project funded by the abrdn Charitable Foundation, the Aage V.~Jensen Naturfond grant no.~N2024-0013, the European Union's Horizon Europe Research and Innovation programme under grant agreement no.~101060639 (MAMBO), and a Discovery Grant by NSERC. The computing resources were provided by Mila~-~Quebec AI Institute, the Digital Research Alliance of Canada, and NVIDIA Corporation.   

A.J., F.C., and M.J.B.~led dataset creation and experiments. J.S.C. and L.P.~contributed to machine learning pipeline development. N.P., F.H, J.R., M.B., M.S., J.F., A.A., Y.L., E.N., F.P.P, A.C.Z., J.A.R.S., and J.G.~performed insect identification for the AMI-Traps dataset. T.A, K.B., A.G.S., M.B., Y.B., K.P.M, D.~Roy, T.T.H., M.L., and D.~Rolnick guided and advised the project.

%
%
\bibliographystyle{splncs04}
\bibliography{egbib}
\clearpage
\setcounter{page}{1}

\renewcommand{\thefigure}{\thesection.\arabic{figure}}
\renewcommand{\thetable}{\thesection.\arabic{table}}

\appendix


\section{Curating the AMI-GBIF Dataset}\label{sec:cleaning_steps_ami_gbif}

In this section, we provide additional details about how the AMI-GBIF dataset is curated.

\begin{figure}[th]
  \centering
  \begin{subfigure}{0.48\linewidth}
    \includegraphics[height=4cm]{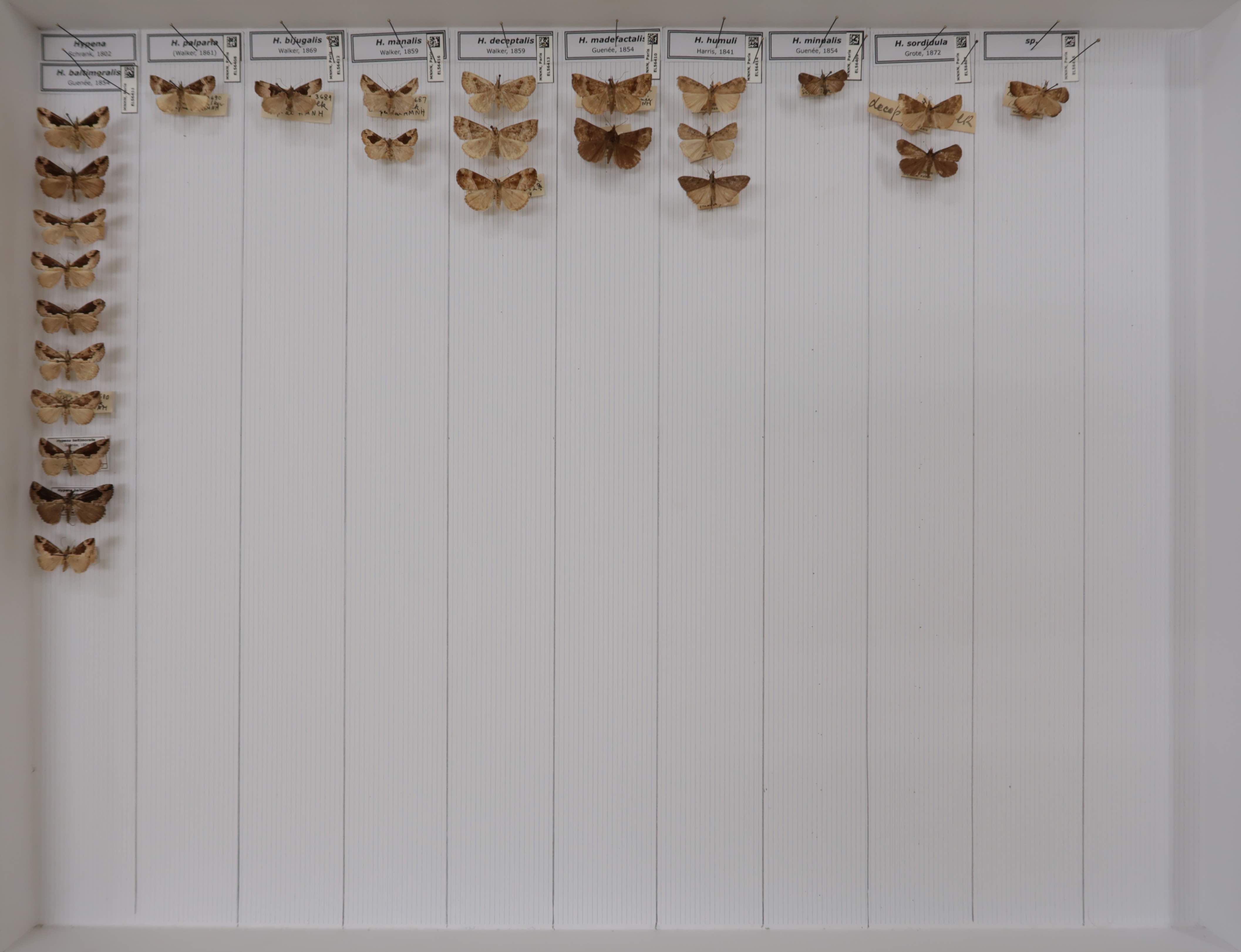}
    \caption{Duplicate URL}
    \label{fig:gbif_duplicate}
  \end{subfigure}
  \hfill
  \begin{subfigure}{0.48\linewidth}
    \includegraphics[height=4cm]{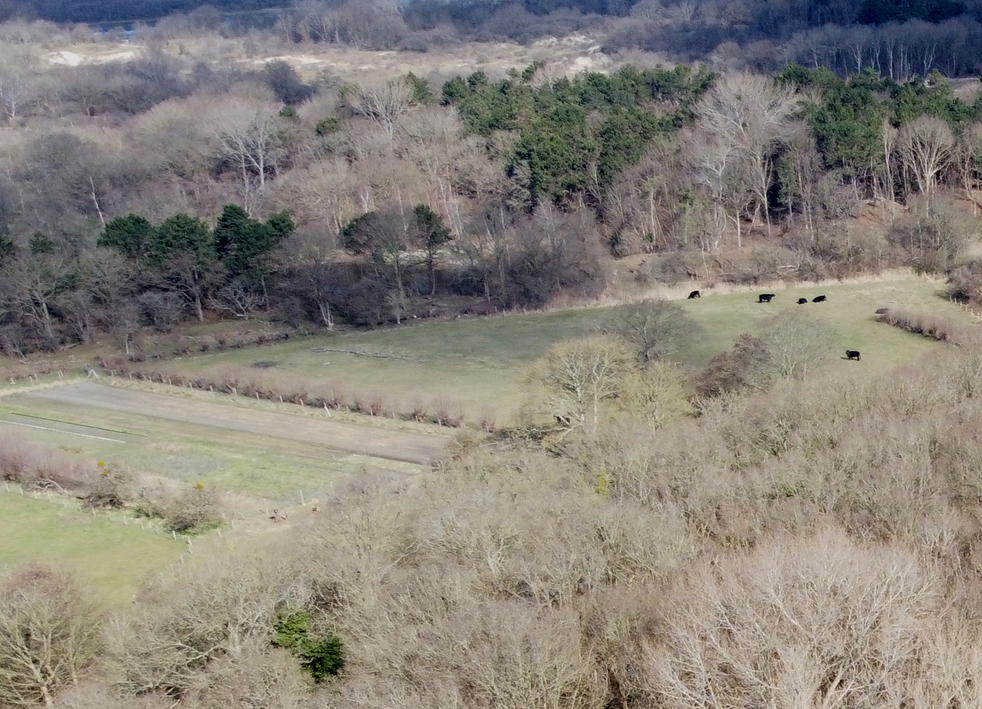}
    \caption{Duplicate URL (placeholder)}
    \label{fig:gbif_duplicate_placeholder}
  \end{subfigure}
  
  \begin{subfigure}{0.37\linewidth}
    \includegraphics[height=1.5cm]{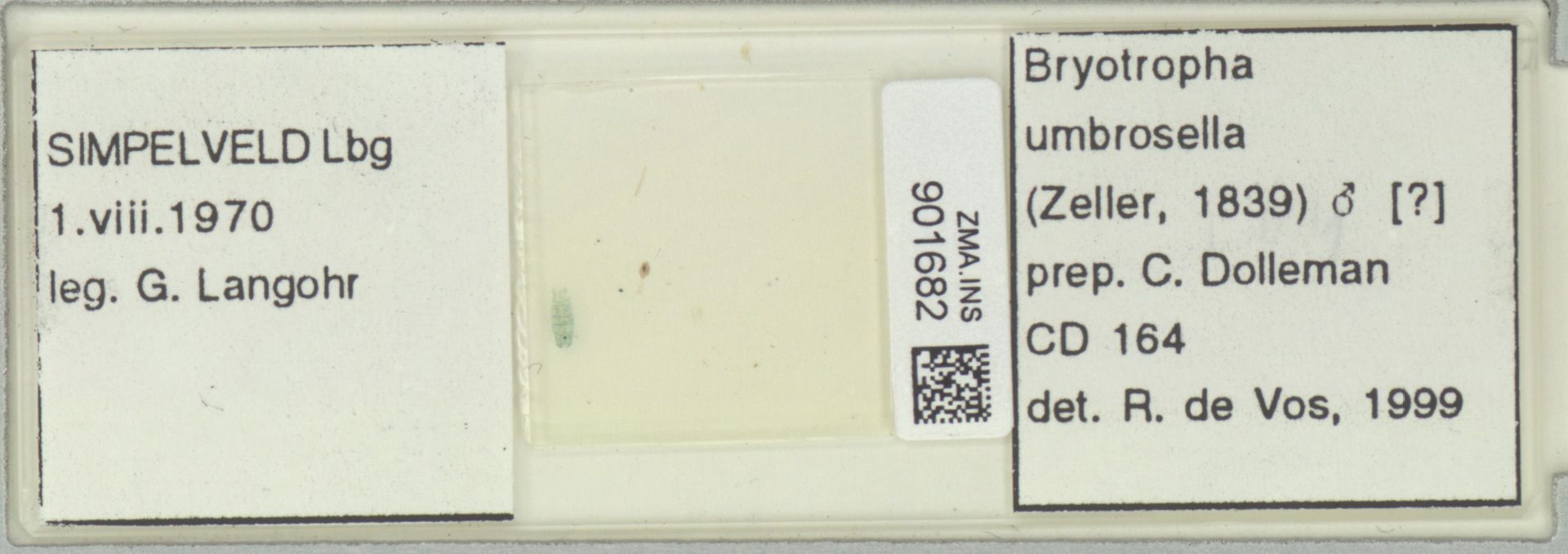}
    \caption{Body parts}
    \label{fig:gbif_body}
  \end{subfigure}
   \hfill
  \begin{subfigure}{0.2\linewidth}
    \includegraphics[height=1.5cm]{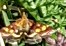}
    \caption{Thumbnail}
    \label{fig:gbif_thumb}
  \end{subfigure}
   \hfill
  \begin{subfigure}{0.3\linewidth}
    \includegraphics[height=1.5cm]{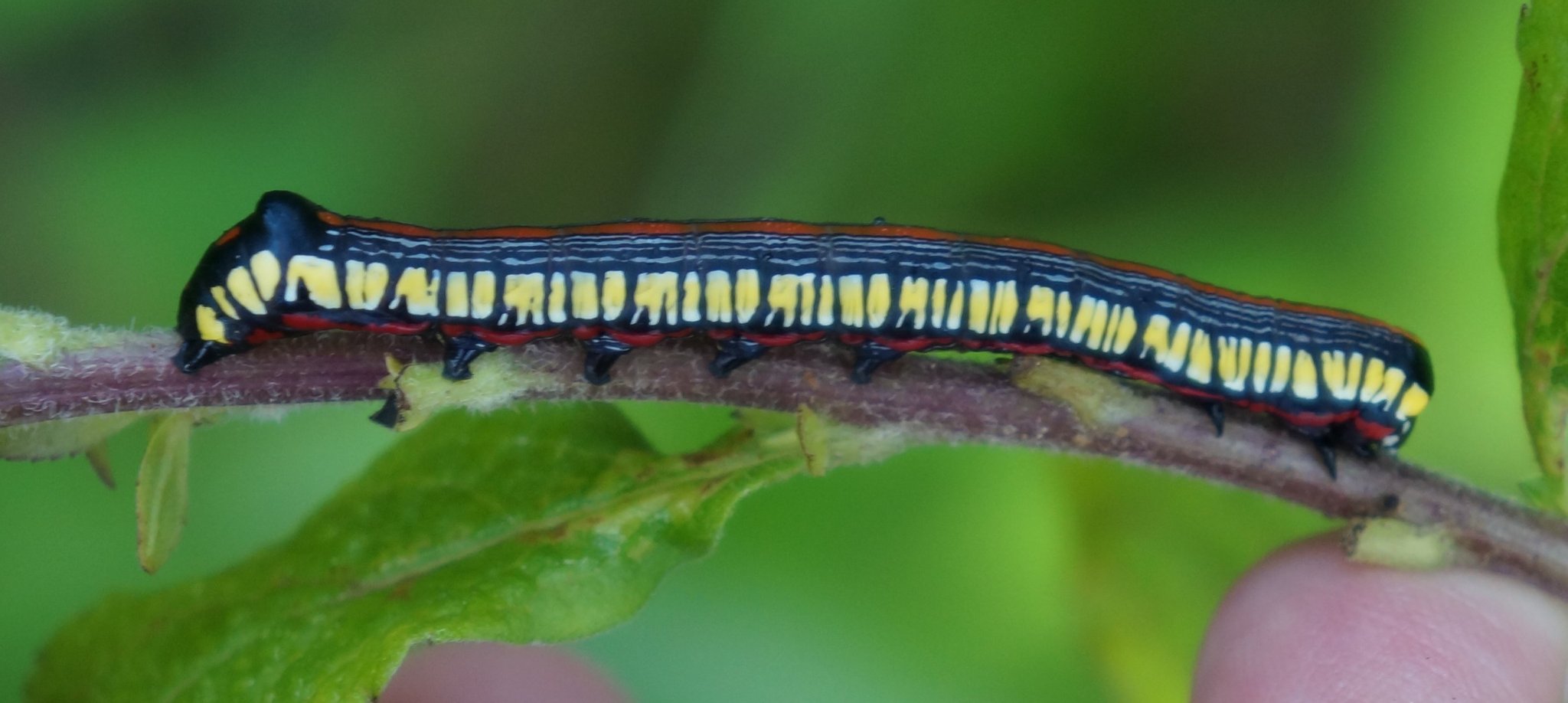}
    \caption{Non-adult}
    \label{fig:gbif_nonadult}
  \end{subfigure}
  \caption{Examples of pictures removed during the dataset cleaning: (a) Contains multiple specimens referenced by multiple observations, (b) a placeholder for more than 100,000 observations, (c) an unsuitable picture containing only a moth body part, (d) a thumbnail, and (e) an animal in the larval life stage.}
  \label{fig:gbif_removed_images}
\end{figure}

\begin{figure}[h]
    \centering
    \includegraphics[width=\textwidth]{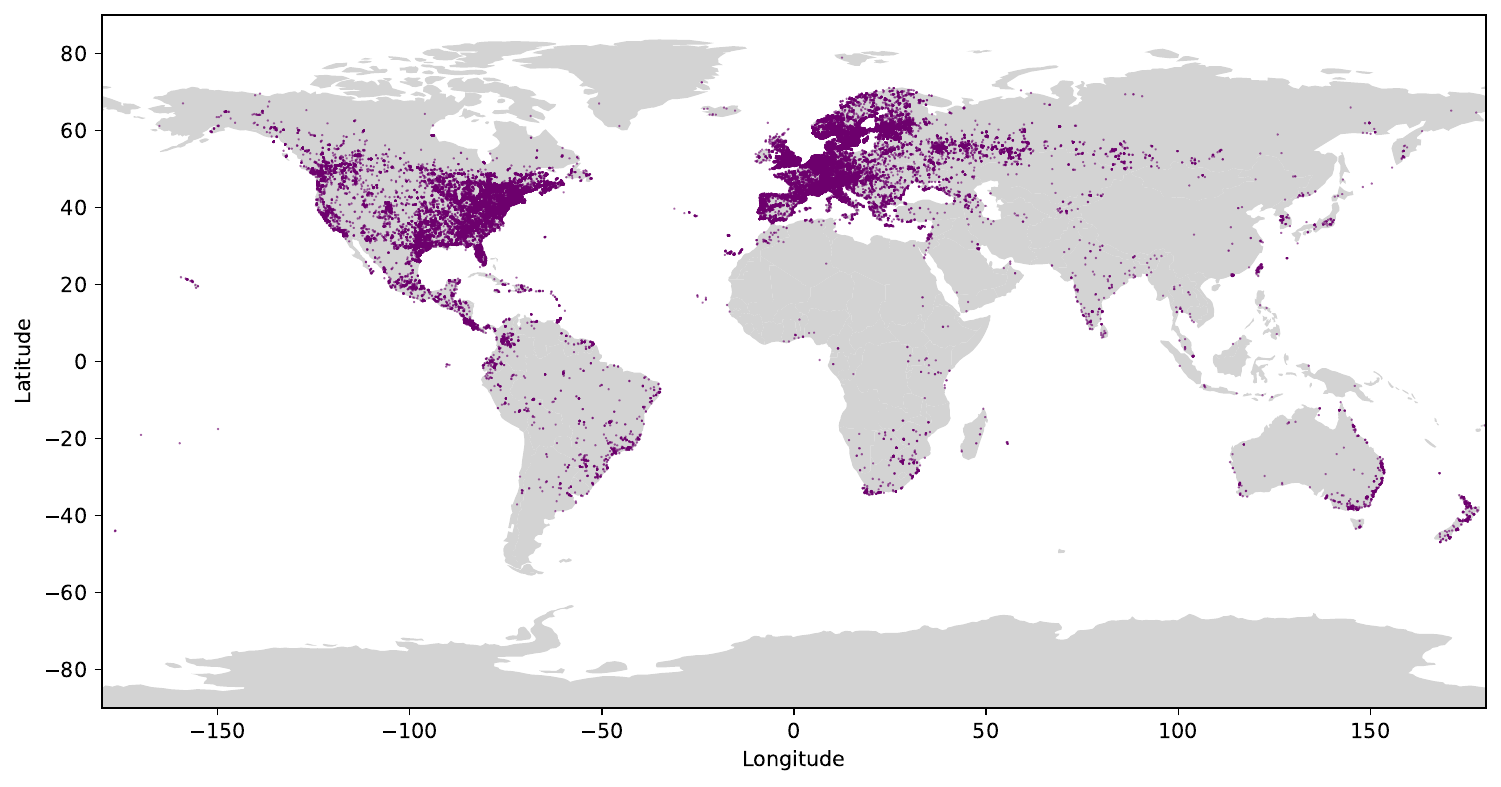}
    \caption{Distribution of geographic locations for images in AMI-GBIF. Since we consider species from the NE-America, C-America, and W-Europe regions, observations are concentrated in these regions. However, some observations fall elsewhere, reflecting the fact that some species present in a region of interest are also found elsewhere; all global observations of that species are drawn upon in creating AMI-GBIF.}
    \label{fig:gbif_global_dist}
\end{figure}

To ensure the quality control of the AMI-GBIF dataset, we remove images affected by issues identified as problematic for training our models: corrupted files, duplicate URLs, images from specific sources, and thumbnails. As detailed below, we also remove images containing non-adult individuals. \autoref{fig:gbif_removed_images} shows some examples of pictures removed during our cleaning steps. \autoref{table:removed_sources} lists the sources we found containing a high percentage of unsuitable images for training models. Though species are selected based on their presence in the NE-America, C-America, and W-Europe regions, we include all GBIF observations of these species from anywhere on Earth, without any geographic filter. \Cref{fig:gbif_global_dist} shows the distribution around the globe of the observations used for the training set of AMI-GBIF. 

\textbf{Life Stage Classifier}: For images without a GBIF life stage tag, we design a life stage classifier to predict whether the picture contains an adult individual or not. In order to train a binary classifier for adult/non-adult moths, we prepare a dataset containing 647,622 images with the life stage tag from the order Lepidoptera. Of this total, 46.35\% are adults, 46.35\% are larvae, 4.7\% are pupae, and 2.6\% are eggs. We group all non-adults into a single category. Next, we split the dataset into training/validation/testing sets, keeping the proportion of 70\%/10\%/20\%, respectively. Then, we train an EfficientNetV2-B3 model with an input size of 300 pixels for 10 epochs, achieving an accuracy of 98.67\% on the validation set.

\begin{table}
\centering
\caption{List of sources from which images were removed from AMI-GBIF due to a high percentage of unsuitable images for training models. Examples include pictures of body parts and images with descriptive text only.}
\label{table:removed_sources}
\begin{tabular}{|cc|}
\hline
\multicolumn{1}{|c}{datasetKey}       & \multicolumn{1}{c|}{Source}      \\ \hline
f3130a8a-4508-42b4-9737-fbda77748438 & Naturalis Biodiversity Center   \\
4bfac3ea-8763-4f4b-a71a-76a6f5f243d3 & Museum of Comparative Zoology   \\
7e380070-f762-11e1-a439-00145eb45e9a & Natural History Museum (London) \\ \hline
\end{tabular}
\end{table}


\section{Creating the AMI-Traps Dataset}\label{sec:ami-traps-construction-appendix}
This section discusses additional details on the camera trap deployments and the custom model developed for insect detection in the camera trap images.

\textbf{Data Collection.} 
The majority of the source images used to create the AMI-Traps dataset were captured by a type of autonomous monitoring system designed by the UK Center for Hydrology and Ecology (though the hardware was modified in different locations to suit local needs and constraints).  The cameras are programmed to run throughout the night, typically every-other day from 22:00 to 05:00. Insects are attracted by a ultraviolet light and the camera is triggered by motion, as well as on a time interval of 10 minutes.  \Cref{tab:source_images} shows the number of deployments, operational nights, and images annotated for each region.

\begin{table}[tbh]
    \centering
    \caption{Sources of images captured by the AMI camera traps and used to curate the AMI-Traps dataset.}
    \begin{tabular}{|l|c|c|c|} 
        \hline 
        Region & Cameras & Nights & Images Annotated\\
        \hline 
        \multicolumn{1}{|l|}{Denmark} & 11 & 155 & 892\\
        \multicolumn{1}{|l|}{Panama} & 2 & 3 & 122\\
        \multicolumn{1}{|l|}{Quebec} & 2 & 159 & 533\\
        \multicolumn{1}{|l|}{UK} & 4 & 38 & 446\\
        \multicolumn{1}{|l|}{Vermont} & 3 & 130 & 900\\
        \hline
        \multicolumn{1}{|l|}{Total} & 22 & 485 & 2893\\
        \hline
    \end{tabular}    
    \label{tab:source_images}
\end{table}

\textbf{Insect Detector.} In this paper, we do not assess object detection as a benchmark task; instead it is used as an initial step for annotating the images, where an object detection model suggests bounding boxes that are then corrected by the human annotators. While the AMI-Traps images have a uniform background, it is still not a simple object detection task. Due to the challenges described in \cref{fig:challenges-ami-traps}, even foundation models for object detection like the Segment Anything Model (SAM)~\cite{kirillov2023segment} make errors on this data. These include a strong tendency to crop out legs and antenna, failure to detect smaller moths, and low inference speed. Hence, we developed a custom model.

The lack of abundant labeled data is the biggest challenge in training an object detector for insect camera traps. We use a weak annotation system leveraging SAM to generate a synthetic training dataset to minimize manual annotation. First, we use SAM to segment nearly 4k insect crops from 300 trap images from five trap deployments. Second, we review the crops to remove undesirable images, which gives us 2600 clean crops. Each crop review is much faster than drawing a bounding box. Third, the crops are randomly pasted on empty background images with simple augmentations (flips and rotations) to create a large simulated labeled dataset of 5k images. We train two versions of Faster R-CNN models~\cite{ren2015faster} on this synthetic dataset: a slow model (ResNet-50-FPN backbone) and a fast model (MobileNetV3-Large-FPN backbone). The latter is six times faster than the former on a CPU while having a near-similar accuracy. We use the MobileNet version to generate bounding box suggestions on the Label Studio platform which are then corrected by our expert annotators before being identified.


\section{Dataset}\label{sec:dataset-appendix}
\begin{figure}[t!]
     \centering
     \begin{subfigure}{0.48\textwidth}
         \centering
         \includegraphics[width=\textwidth]{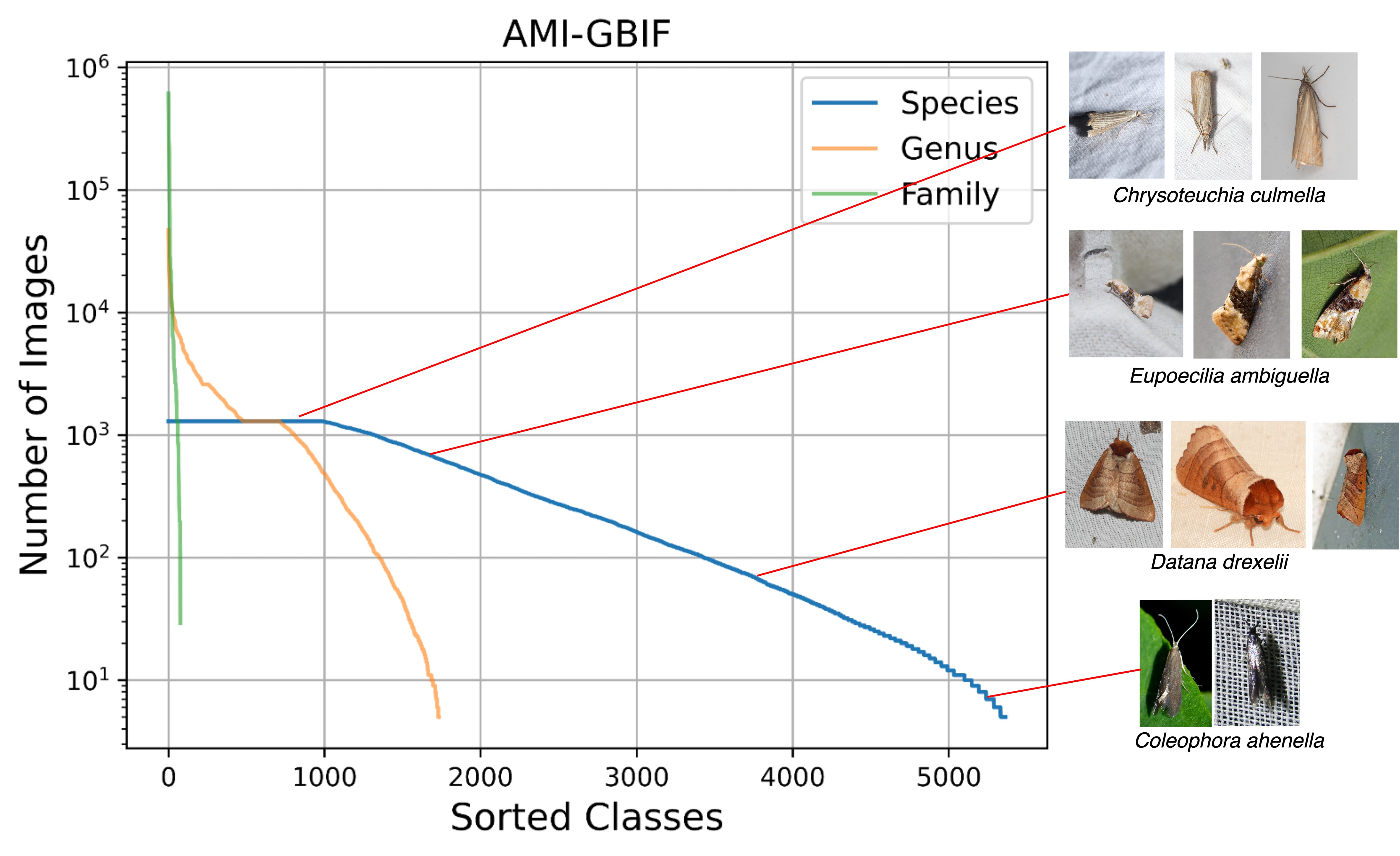}
     \end{subfigure}
     \begin{subfigure}{0.48\textwidth}
         \centering
         \includegraphics[width=\textwidth]{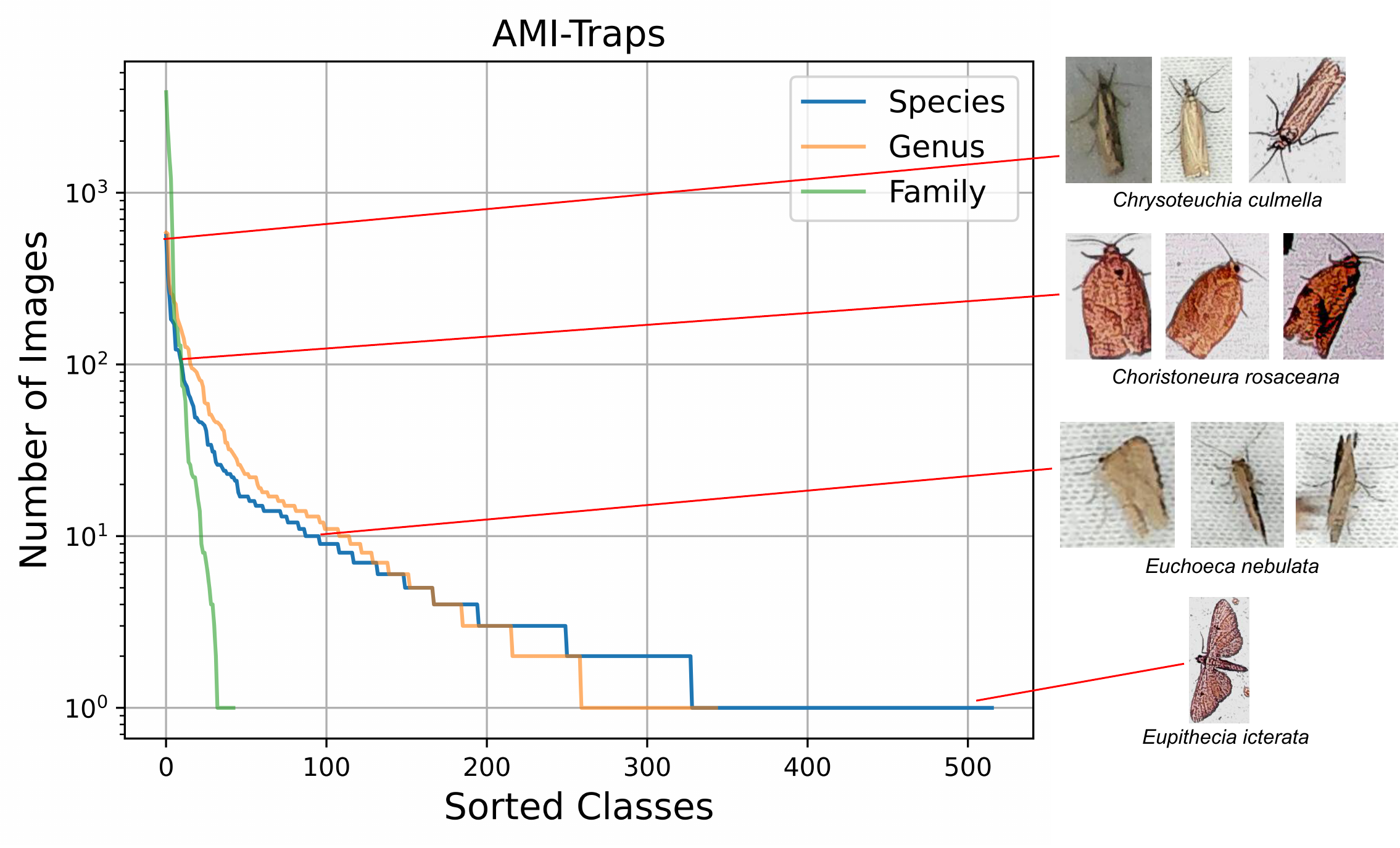}
     \end{subfigure}     
        \caption{Long-tailed distribution of the data in the AMI-GBIF (left) and AMI-Traps (right) datasets. The data is skewed both at the species level and also at the genus and family levels.}
        \label{fig:long_tailed_distribution}
\end{figure}

\Cref{fig:long_tailed_distribution} shows the long-tailed nature of the AMI dataset.

\section{Data Augmentation}\label{sec:data-aug}
\begin{figure}[tbh]
    \centering
    \includegraphics[width=\textwidth]{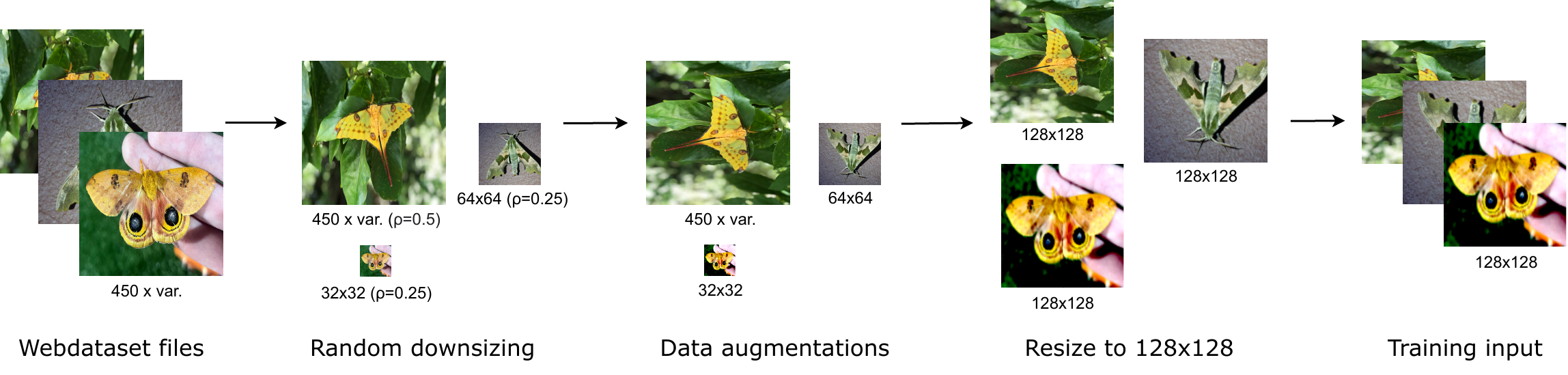}
    \caption{The original AMI-GBIF images are stored in the WebDataset format with the smallest dimension of 450px and the other being variable to preserve the aspect ratio. First, with a probability of 0.25 each, images are downsized to a side length of 64 or 32 pixels, while the rest maintain their original size. Next, additional data augmentation techniques are applied, which include random resize crop,  random horizontal flip, and RandAugment \cite{cubuk2020randaugment}. Finally, the images are resized to 128 pixels before being used as training data.} 
    \label{fig:mixed-resolution}
\end{figure}

\Cref{fig:mixed-resolution} shows the data augmentation workflow in our model training pipeline. The MixRes data augmentation approach helps models generalize from AMI-GBIF training images (which are typically high resolution) to AMI-Traps test images (which are typically very low resolution).

\section{Additional Experiments}\label{sec:extra_exp}

In this section, we provide additional experimental results.

\textbf{Binary Classification Ablation.} \Cref{tab:binary_eval_reshaping} shows results for an ablation study where the binary classification of the AMI-Traps images is performed with and without square padding before model prediction. The padding leads to an increase in precision.

\textbf{Long-Tailed Results for AMI-GBIF}. The AMI-GBIF dataset is a valuable benchmark dataset for fine-grained and long-tailed classification tasks, providing a real-world scenario for evaluating new methods for these kinds of problems. We present additional results to demonstrate the performance of our baseline models regarding the long-tailed distribution. Alongside the general accuracy, we assess accuracy within buckets based on the number $N$ of training instances per category: many-shot ($N > 100$), medium-shot ($20 \leq N \leq 100$), and few-shot ($N < 20$). The results (see \autoref{tab:ami-gbif-long-tail-results}) show that ConvNeXt-B achieves the best performance in all regimes, especially for the few-shot classes, with around 8\% higher accuracy than other families of architectures and more than 2.5\% higher than its tiny version, ConvNeXt-T.

\textbf{Pre-Training Ablation.} \Cref{tab:pretraining-abla} shows an ablation study to check the influence of pre-training on fine-grained classification accuracy. It is evident that using ImageNet weights and pre-training on all AMI-GBIF data followed by region-specific fine-tuning achieves the best performance. Training from scratch also shows good results but requires much longer epochs.

\begin{figure}[t!]
     \centering
         \begin{subfigure}{0.47\textwidth}
         \centering
         \includegraphics[width=\textwidth]{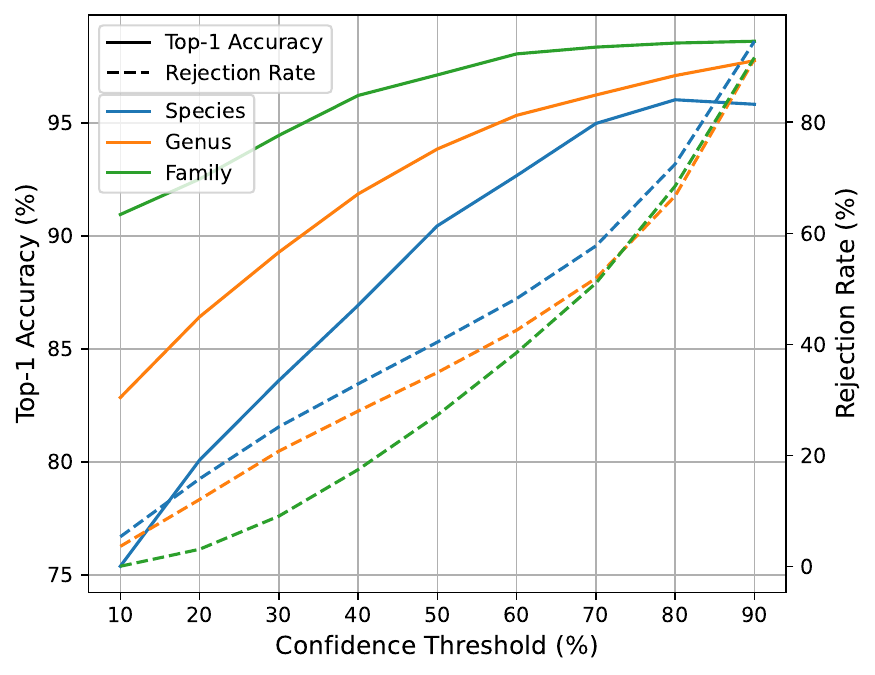}
     \end{subfigure}
     \begin{subfigure}{0.49\textwidth}
         \centering
         \includegraphics[width=\textwidth]{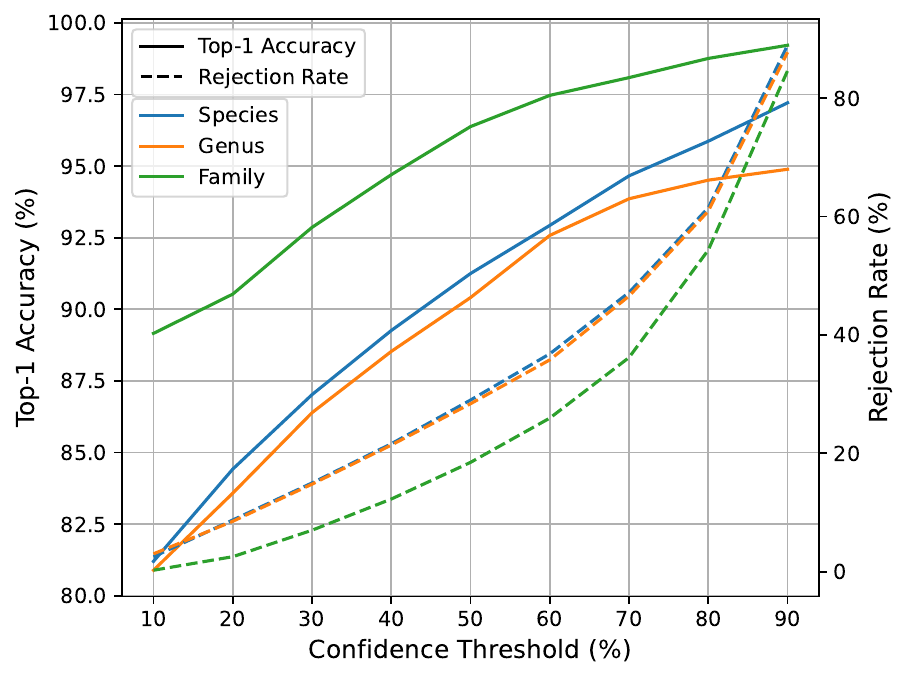}
     \end{subfigure}     
        \caption{The left y-axis compares the top-1 accuracy as a function of model (ResNet50) confidence, while the right y-axis shows the rejection rate (ratio of rejected samples to total samples) for the corresponding thresholds for the NE-America (left) and W-Europe regions (right). At a confidence of 50\%, the model has over 90\% accuracy at all taxonomic levels. This analysis is skipped for the C-America which has sparse taxonomic coverage in the AMI-T dataset.}
        \label{fig:performance_top1_accuracy}
\end{figure}

\textbf{Analysis of Accuracy v/s Confidence}: One important point for practical use cases is how accuracy changes if predictions are only accepted above a certain "confidence threshold". Thresholding predictions can increase the reliability of model outputs for downstream applications, though it comes with a trade-off of rejecting a certain fraction of predictions. \Cref{fig:performance_top1_accuracy} shows the curves of the accuracy and rejection rate as functions of the confidence threshold.

\begin{table}[tb]
  \caption{Comparison of the binary classification accuracy when the insect crops in the AMI-Traps dataset are not square padded (w/o pdg.) and square padded (pdg.) with black pixels before model prediction.}
  \label{tab:binary_eval_reshaping}
  \centering
  \resizebox{\textwidth}{!}{%
  \begin{tabular}{|l|*{3}{P{1.8cm}P{1.8cm}|}}
  \hline
  & \multicolumn{2}{c|}{All Crops} & \multicolumn{2}{c|}{Crops $>$ 100px} & \multicolumn{2}{c|}{Crops $>$ 150px} \\
  & w/o pdg. & pdg. & w/o pdg. & pdg. & w/o pdg. & pdg. \\
  \hline
  Precision & 66.21$\pm$2.04 & 68.31$\pm$1.70 & 78.10$\pm$1.46 & 79.46$\pm$1.43 & 87.96$\pm$0.84 & 89.83$\pm$0.43\\
  Recall & 94.00$\pm$0.72 & 95.03$\pm$0.39 & 96.28$\pm$0.24 & 96.99$\pm$0.20 & 97.27$\pm$0.20 & 97.48$\pm$0.12 \\
  \hline
  \end{tabular}
  }
\end{table}

\begin{table}[h!]
\centering
\caption{Micro-averaged long-tailed accuracy results for AMI-GBIF. ConvNeXt-B achieved the best performance for all regions, including a considerably higher accuracy for the medium- and few-shot species.}
\label{tab:ami-gbif-long-tail-results}
\begin{tabular}{|l|l|P{1.8cm}P{1.8cm}P{1.8cm}P{1.8cm}|}
\hline
\multicolumn{1}{|c|}{\multirow{2}{*}{Region}} &  \multicolumn{1}{c|}{\multirow{2}{*}{Model}} &
  \multicolumn{4}{c|}{Species} \\
 &
   &
  \multicolumn{1}{c}{All} &
  \multicolumn{1}{c}{Many} &
  \multicolumn{1}{c}{Medium} &
  \multicolumn{1}{c|}{Few} \\ \hline
\multirow{5}{*}{NE-America} & ResNet50         &     87.56$\pm$0.12     &  88.84$\pm$0.09   &  66.99$\pm$0.57  & 34.03$\pm$0.68  \\
                            & MBNetV3Large &   83.73$\pm$0.17      &  85.12$\pm$0.19   &  61.48$\pm$0.47  &  26.10$\pm$1.58 \\
                            & ConvNeXt-T       &      89.81$\pm$0.05     &  90.91$\pm$0.05   &  72.57$\pm$0.21  &  40.74$\pm$1.35 \\
                            & ConvNeXt-B       &      \textbf{90.45$\pm$0.03}      &  \textbf{91.47$\pm$0.04}   & \textbf{74.59$\pm$0.20}   & \textbf{45.81$\pm$1.30}  \\
                            & ViT-B/16         &     86.39$\pm$0.05      &  87.58$\pm$0.05   &  67.64$\pm$0.23  & 35.27$\pm$1.10  \\ \hline
\multirow{5}{*}{W-Europe}   & ResNet50         &    86.45$\pm$0.08        &  87.62$\pm$0.08   &  58.74$\pm$0.36  & 27.23$\pm$0.99  \\
                            & MBNetV3Large &          82.20$\pm$0.05        &  83.46$\pm$0.07   & 52.15$\pm$0.51  & 18.05$\pm$0.90  \\
                            & ConvNeXt-T       &     88.82$\pm$0.03        &   89.84$\pm$0.04   & 64.85$\pm$0.49   & 33.30$\pm$0.45 \\
                            & ConvNeXt-B       &      \textbf{89.35$\pm$0.01}       &   \textbf{90.31$\pm$0.01}  &  \textbf{66.67$\pm$0.33}   &  \textbf{38.39$\pm$1.95}  \\
                            & ViT-B/16         &      85.45$\pm$0.03       &  86.56$\pm$0.03   & 59.14$\pm$0.13  &  27.56$\pm$0.85 \\ \hline
\multirow{5}{*}{C-America}  & ResNet50         &   87.46$\pm$0.23       & 90.90$\pm$0.13    & 79.64$\pm$0.47   &  49.00$\pm$1.84  \\
                            & MBNetV3Large &     83.68$\pm$0.24       &  87.82$\pm$0.19   & 72.72$\pm$0.25   & 42.01$\pm$1.62  \\
                            & ConvNeXt-T       &     87.65$\pm$0.11        &  90.54$\pm$0.11   &  81.45$\pm$0.13   & 54.32$\pm$1.27   \\
                            & ConvNeXt-B       &     \textbf{88.62$\pm$0.09}       &  \textbf{91.23$\pm$0.14}   &  \textbf{83.56$\pm$0.19}   &  \textbf{56.91$\pm$0.59}  \\
                            & ViT-B/16   .      &      79.80$\pm$0.26      &  83.77$\pm$0.32   &  69.50$\pm$0.53   &  39.21$\pm$0.91  \\ \hline
\end{tabular}
\end{table}

\begin{table}[]
\centering
\caption{Ablation study results for different choices of pre-training. PT and FT stands  for pre-training and fine-tuning respectively. When training from scratch, we use a schedule of 90 epochs and an initial learning rate of 0.01. As the differences in results may be due to the number of epochs, we keep the total epochs consistent (30 or 90) when pretraining on all AMI-GBIF data and then fine-tuning on regional checklists. The model architecture is ResNet50 with ImageNet weights loaded from the timm library, wherever applicable. The results are species accuracy for the NE-America region.}
\label{tab:pretraining-abla}
\begin{tabular}{|P{1.4cm}P{1.6cm}P{1.4cm}|P{1.4cm}|P{1.8cm}P{1.8cm}P{1.8cm}|}
\hline
ImageNet & AMI-GBIF & PT Epochs & FT Epochs & AMI-GBIF$\ddag$ & AMI-Traps (Micro) & AMI-Traps (Macro) \\ \hline
\checkmark      & \crossmark       & -      & 30        &    89.39$\pm$0.14     &       71.86$\pm$1.02  & 64.96$\pm$2.00  \\
\crossmark       & \crossmark       & -      & 90        &   88.85$\pm$0.04       &      74.80$\pm$0.82 & 67.45$\pm$0.81     \\
\crossmark       & \checkmark      & 60     & 30        &    79.44$\pm$2.50      &       69.07$\pm$0.76 & 62.58$\pm$1.16  \\
\checkmark      & \checkmark      & 60     & 30        &     90.28$\pm$0.78     &         \textbf{75.06$\pm$0.42} & \textbf{67.34$\pm$0.71}  \\
\checkmark      & \checkmark      & 20     & 10        &     \textbf{90.60$\pm$0.05}     &         73.57$\pm$0.52 & 66.16$\pm$0.51    \\ \hline
\end{tabular}
\end{table}

\end{document}